\documentclass[conference]{IEEEtran}
\usepackage{graphicx}
\usepackage{hyperref}
\usepackage{subfig}
\usepackage{multirow}
\usepackage{amsmath}
\usepackage{lmodern}
\usepackage{url}
\usepackage{booktabs}
\usepackage{algorithm,algorithmic}

\def\BibTeX{{\rm B\kern-.05em{\sc i\kern-.025em b}\kern-.08em
 T\kern-.1667em\lower.7ex\hbox{E}\kern-.125emX}}
\begin{document}
\title{SSRepL-ADHD: Adaptive Complex Representation Learning Framework for ADHD Detection from Visual Attention Tasks}
\author{\IEEEauthorblockN{Abdul Rehman\textsuperscript{1}, Ilona Heldal\textsuperscript{1}, and Jerry Chun-Wei Lin\textsuperscript{1}
}
\IEEEauthorblockA{
\textsuperscript{1}Department of Computer Science, Electrical Engineering and Mathematical Sciences\\
Western Norway University of Applied Sciences\\ Bergen, Norway\\
Email: {\{arj, ilona.heldal, jerry.chun$-$wei.lin\}}@hvl.no
}
}
\maketitle
\begin{abstract}
Self Supervised Representation Learning (SSRepL) can capture meaningful and robust representations of the Attention Deficit Hyperactivity Disorder (ADHD) data and have the potential to improve the model's performance on also downstream different types of Neurodevelopmental disorder (NDD) detection. In this paper, a novel SSRepL and Transfer Learning (TL)-based framework that incorporates a Long Short-Term Memory (LSTM) and a Gated Recurrent Units (GRU) model is proposed to detect children with potential symptoms of ADHD. This model uses Electroencephalogram (EEG) signals extracted during visual attention tasks to accurately detect ADHD by preprocessing EEG signal quality through normalization, filtering, and data balancing. For the experimental analysis, we use three different models: 1) SSRepL and TL-based LSTM-GRU model named as \textit{SSRepL-ADHD}, which integrates LSTM and GRU layers to capture temporal dependencies in the data, 2) lightweight SSRepL-based DNN model (\textit{LSSRepL-DNN}), and 3) Random Forest (RF). In the study, these models are thoroughly evaluated using well-known performance metrics (i.e., accuracy, precision, recall, and F1-score). The results show that the proposed \textit{SSRepL-ADHD} model achieves the maximum accuracy of \texttt{81.11\%} while admitting the difficulties associated with dataset imbalance and feature selection. 
\end{abstract}


\IEEEpeerreviewmaketitle

\section{Introduction}
Self Supervised Representation Learning (SSRepL) \cite{ericsson2022self} captures useful representations of the input data that can potentially improve the model's performance on downstream tasks. ADHD is a prevalent neurodevelopmental condition that affects both children and adults, with 5.6\% for teenagers aged 12-18 and 7\% for those under 12, globally (according to work examined by Salari et al. \cite{salari2023global}). Symptoms such as impulsive actions, hyperactivity and lack of concentration are indicative of the disorder. In the literature, EEG has been used to diagnose ADHD \cite{taghibeyglou2022detection,motamed2022recognition} since several EEG features have been used as markers for the detection of ADHD. Researchers have also investigated connectivity features and event-related potentials extracted from EEG data to understand better and classify ADHD \cite{alves2023identifying}. An effective, information-based, functional and nonlinear method to study EEG channel synchronization is EEG connectivity. Effective connectivity uses Granger causality and partial directional coherence, while functional connectivity uses methods such as coherence and cross-correlation. The two main types of nonlinear connectivity metrics are generalized synchronization and phase synchronization \cite{sakkalis2008assessment}. EEG functional connectivity is useful as a marker for developmental brain disorders such as ADHD. Key connectivity metrics in this setting include EEG coherence and the weighted phase lag index, which have successfully differentiated between individuals with and without ADHD \cite{debnath2021investigating, furlong2021resting, kiiski2020functional}.

Several studies have been conducted on neurological disorders diagnosed using EEG connectivity, not only ADHD, epilepsy, Parkinson's disease and autism \cite{briels2020reproducibility, van2019network, bovckova2019impairment, bovckova2019impairment,ahmadi2021computer}. The complex dynamics of brain function can be better understood using different connectivity approaches that help to identify and understand various neurological disorders. Machine learning techniques have been used extensively in the classification of medical diseases and disorders, such as ADHD, in recent years \cite{taghibeyglou2022detection,wang2022attention}. However, as far as we know, most of the existing studies have focused on the use of EEG to classify ADHD. However, no generalizable and pre-trained approach can classify almost all downstream tasks. With the above limitations in mind, this paper makes the following contributions.

\subsection{Contribution}
\begin{itemize}
\item In this paper, a novel framework using SSRepL and TL-based hybrid LSTM-GRU models named \textit{SSRepL-ADHD} is proposed, which integrates LSTM and GRU layers to capture complex patterns in the data to recognize ADHD and provide a pre-trained model for downstream tasks. First, the EEG signals are normalized and filtered, and the data are equalized. Two models are used for experimental analysis and comparison: 1) a lightweight SSRepL-based LSTM model (LSSRepL-DNN) and 2) an RF model. In addition, it offers the potential of transfer learning for practical implementation in real-world clinical situations where data is limited, more training time and computational resources are needed, or different but related tasks are required.

 \item  This study provides a comprehensive evaluation method in which the performance of the models is evaluated based on accuracy, precision, recall, F1 score and confusion matrix. The results show that each model performs well, with the SSRepL-ADHD model achieving the maximum accuracy of \texttt{81.11\%}. This in-depth analysis provides a comprehensive understanding of the model's strengths and limitations and ensures a thorough evaluation of its predictive capabilities. We plan to make this pre-trained model public for further ADHD research.
\end{itemize}

\subsection{Research Organization}
This paper is organized as follows. The literature review in Section \ref{lr} lays the groundwork by reviewing previous work. Section \ref{datasetselection} describes the dataset, emphasizing its features and significance. Section \ref{pm} describes the suggested technique in depth, including data preparation, preprocessing, and the novel transfer learning model based on EEG signals. Section \ref{ev} presents empirical findings and analysis, as well as evaluation metrics. Finally, Section \ref{conclusion} wraps up the work by making future recommendations. 

\section{Literature Review} \label{lr}
The identification of ADHD is currently the focus of numerous studies \cite{cortese2022half,loh2022automated,galvez2022therapeutic,faraone2021world}. For example, several research groups have started to develop advanced deep learning and machine learning algorithms based on ADHD data \cite{wang2022attention,hernandez2023machine}. It is worth noting that these algorithms have been studied to improve performance in the diagnosis of ADHD. While ADHD cannot be treated, early detection can contribute to how these ADHD individuals function in society \cite{carpentier2012adhd}. Researchers are trying to identify risk factors to reduce the number of children diagnosed with ADHD. One study found a strong correlation between ADHD and hereditary variables \cite{faraone2021world}. In younger children, around 75\% of the risk of ADHD is caused by genetic factors \cite{faraone2021world}. Brain damage and alcohol or tobacco use during pregnancy and early birth are among the risk factors for ADHD, along with genetics. Several other factors have been linked to ADHD in children \cite{claussen2022all}. These include gender, age, race, asthma, anxiety, depression, obesity, smoking and socioeconomic status. Machine learning (ML) models can be used instead of classical approaches for prediction. The mental health \cite{kessler2019role,zea2022machine} have all used ML-based models for identification and prediction.

Various machine learning classifiers were used to predict which children would raise ADHD \cite{kim2021can,zhang2021evidence}. Uluyagmur-Ozturk et al. \cite{uluyagmur2016adhd} categorized children as having ASD, ADHD or no diagnosis in a study examining their emotional well-being. Information was obtained on 61 children from Marmara University Medical Hospital. 18 children had autism spectrum disorder, 30 children had ADHD and 13 completely healthy children. To identify the hallmarks of ASD and ADHD, they turned to ReliefF. They used five ML-based algorithms to further categorize the children as to whether they were ASD, ADHD or healthy. The results show an 80\% success rate in classifying children as healthy with ASD or ADHD.

In another study, a Continuous Performance Test (CPT) was used to diagnose children with ADHD \cite{slobodin2020machine}. They selected 458 children aged 6 to 12 years. With a mean age of $8.7\pm$1.8 years, 59.0\% of the selected children were male, and 46.51\% of these children struggled with ADHD. Their proposed ML classifier, MOXO, showed an impressive accuracy of 87.0\%.
In addition, Morrow et al. conducted a study of children being treated for ADHD \cite{morrow2020leveraging} in which they retrieved information on 6,630 children aged 3 to 17 years who were diagnosed with ADHD from the National Survey of Children's Health (NSCH), 2016-2017. The children diagnosed with ADHD were, on average, 12.4 years old. They identified the characteristics of children who were receiving treatment for ADHD. They demonstrated that the DeepNet-based classifier achieved the best AUC at 0.72\%.

The application of ML-based classifiers to ADHD diagnosis is still a challenge despite its rapid growth. However, using different ADHD datasets in different countries, deep learning-based classifiers have been used to predict children with ADHD \cite{ahmadi2021computer,kim2021can, zhang2021evidence}. However, there is a need to improve the performance of the models. This study identifies which variables in children are at risk for developing ADHD and proposes a deep learning (DL) classifier that can identify and predict whether a child is healthy or has ADHD. Furthermore, it is possible to provide a generalizable and pre-trained framework that can be used for the classification of almost all downstream tasks in ADHD.

\section{Dataset selection}\label{datasetselection}
The dataset used in this study is publicly available via the IEEE data portal\footnote{https://ieee-dataport.org/open-access/eeg-data-adhd-control-children}. An experienced psychiatrist specializing in the treatment of children and adolescents used DSM-IV criteria to diagnose 60 typically developing children and 60 children with ADHD. For as long as six months, the youngsters with ADHD had received Ritalin. To treat ADHD, Ritalin is prescribed. It alters the levels of certain endogenous chemicals in the brain that are thought to contribute to increased \cite{stein2022psychiatric}. There is no conclusive evidence that Ritalin can affect the distinctness of brain waves in children with ADHD. Two elementary schools were used to select the healthy group, and we included only right-handed students in this study. TABLE \ref{datasett} summarizes the participants' information.

\begin{table}[!ht]
\caption{Participants Information of the database}
\label{datasett}
\centering
\begin{tabular}{@{}|l|l|l|l|l|@{}}
\hline
 & Girls & Boys & Age & Leading Hand \\ \hline
\multicolumn{1}{|l|}{ADHD Children} & \multicolumn{1}{l|}{12} & \multicolumn{1}{l|}{48} & \multicolumn{1}{l|}{7-12} & \multicolumn{1}{l|}{Right Handed} \\ \hline
Healthy Children & 10 & 50 & 7-12 & Right Handed \\ \hline
\end{tabular}
\end{table}

\begin{figure}[!ht]
\centering
\includegraphics[width=\columnwidth]{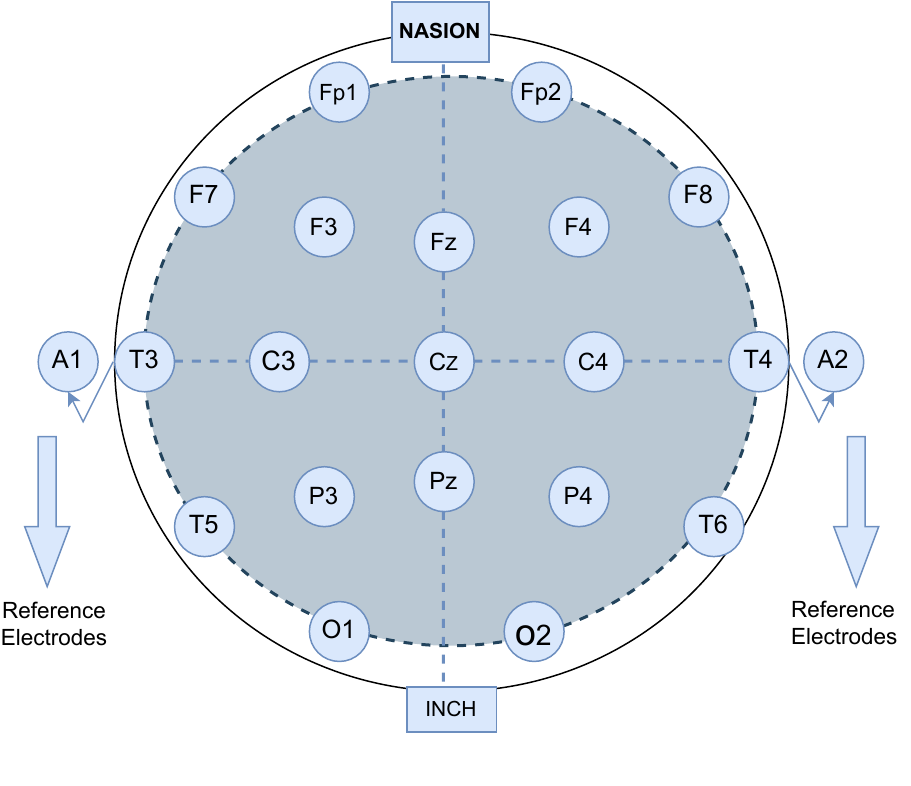}
\caption{The 10-20 methods for electrode placement using reference electrodes A1 and A2.}
\label{signals} 
\end{figure}

In the Psychology and Psychiatry Research Center of Roozbeh Hospital (Tehran, Iran), EEG signals were recorded with a digital instrument (SD-C24, Sholeh Danesh Co., Tehran, Iran) \cite{ekhlasi2021direction}. At a sampling rate of 128 Hz, an electroencephalogram (EEG) with 19 channels (Fz, Cz, Pz, C3, T3, C4, T4, Fp1, Fp2, F3, F4, F7, F8, P3, P4, T5, T6, O1, O2) was recorded according to the 10-20 standard. The earlobe-mounted A1 and A2 electrodes served as the standard. The electrode placements of the worldwide 10-20 system for EEG are shown in Fig. \ref{signals} \cite{alim2023automatic}. 

A visual attention challenge was the basis for the design of the recording protocol. The activity required children to identify 20 pictures featuring various age-appropriate characters, such as different animals. Each picture had a random number of characters ranging from five to sixteen. Each picture was shown immediately after the child's reaction to maintain a constant stimulus throughout the EEG recording. Therefore, the length of the EEG recording is determined by the child's performance. The accuracy of the responses was \cite{ekhlasi2021direction}. The Ethics Committee and the Institutional Review Board of Tehran University of Medical Sciences (TUMS) approved the methods used to collect this dataset \cite{ekhlasi2021direction}. The dataset was collected through a visual attention task that showed age-appropriate and friendly images to children ages 7 to 12. This was done because visual attention is one of the deficiencies in children with ADHD \cite{alim2023automatic}. The balanced dataset showed no measurement bias as the data was collected in two locations and two sessions. Considering these factors and the adaptability in collecting datasets from youngsters, we decided to use this particular dataset. Finally, the dataset contains 2,16,6383 rows and 19 columns. The dataset contains 1,20,7069 rows for ADHD children and 9,59,314 for healthy children.

\section{Proposed Framework} \label{pm}
\begin{figure*}[!ht]
\centering
\includegraphics[width=\textwidth]{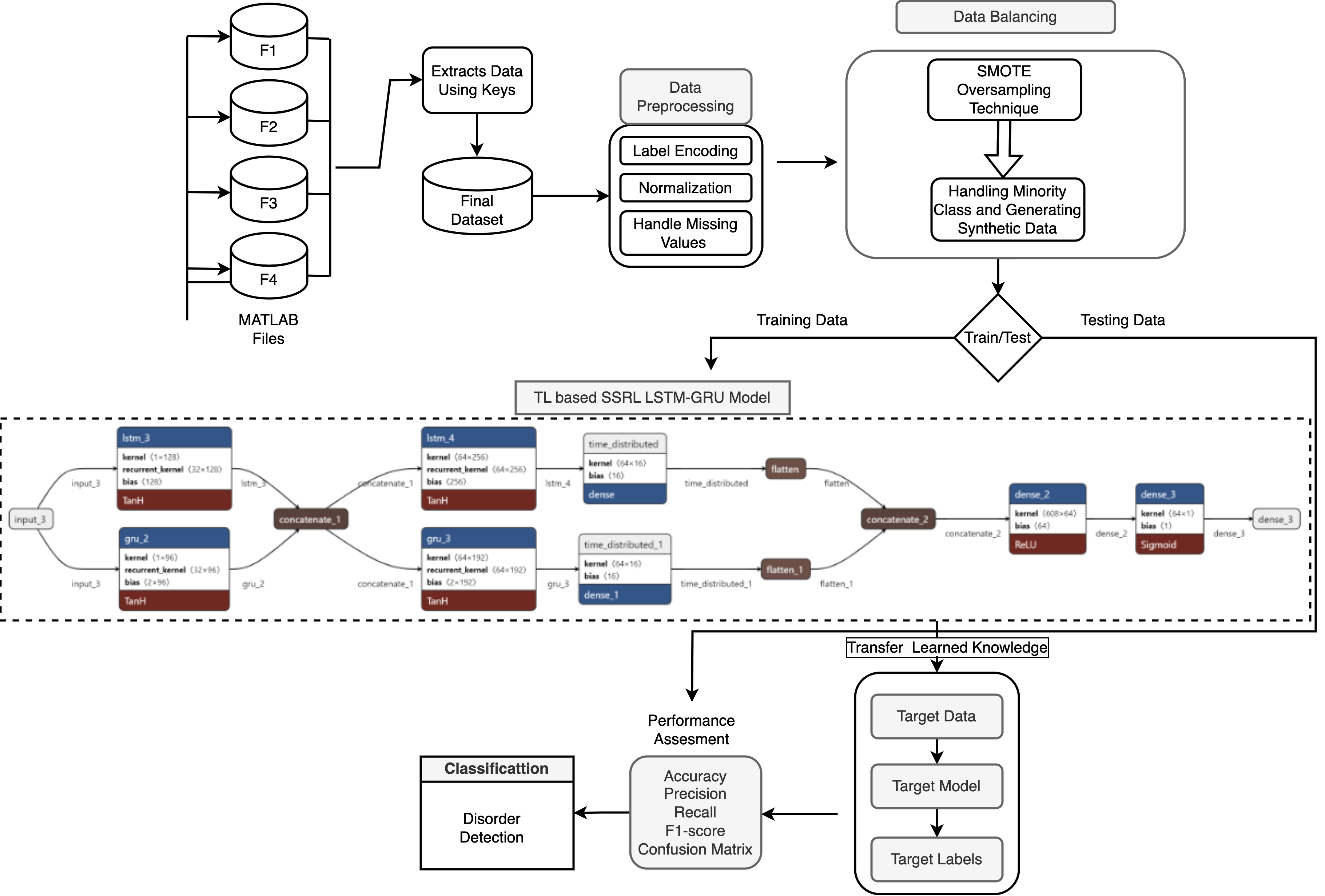}
\caption{Transfer Learned Knowledge based Methodology For ADHD Disease Detection}
\label{proposed} 
\end{figure*}

Fig. \ref{proposed} depicts a comprehensive framework developed for learning representations and identifying ADHD patients using EEG signal analysis. The framework is implemented in several steps, with each step contributing to the refinement and accuracy of the final model. In the preliminary phase, an extensive data preparation process is carried out, including tasks such as data cleaning and removal of unnecessary features. These processes are crucial for assembling a high-quality dataset and building a basis for later model implementation. After data preparation, the necessary pre-processing processes are carried out to extract useful insights from the dataset. Normalization and filtering techniques are used to improve the quality of the EEG signals and enable more effective analysis and interpretation. Data balancing techniques are used to create a balanced data set. This balancing is crucial to ensure that the resulting models provide accurate and unbiased results. The prepared dataset is then subjected to various models, each of which plays an important role in capturing the underlying patterns suggestive of ADHD. In particular, SSRepL and TL approaches will be used to create a framework that utilizes the strengths of the current models to improve the prediction results. A set of evaluation metrics is used to evaluate model performance. These metrics provide a quantifiable measure of the model's accuracy in predicting ADHD and ensure thorough examination. The experiments in this paper used predetermined features, emphasizing the importance of each feature. All selected features were deemed critical to the experiments, emphasizing their importance in achieving the research objectives.

\subsection{Data Pre-Processing}
Three methods are used in the pre-processing phase: Label encoding converts the category data into a numerical format so that the machine/deep learning algorithms can analyze it effectively. Normalization is used to scale numerical features to a constant range, preventing the overwhelming influence of greater magnitude features. In addition, missing value handling tactics are used to deal with incomplete data points using techniques such as imputation or elimination. These pre-processing methods provide a well-structured, representative dataset for effective training of machine/deep learning models. Data balancing resolves class imbalance distribution within a dataset to ensure that machine learning models are not biased towards the dominant class. The Synthetic Minority Over-Sampling Technique (SMOTE) generates synthetic samples for the minority class to achieve a more equal representation. Minority class treatment and synthetic data generation approaches focus on enhancing instances of the minority class to increase its presence in the dataset. 

\begin{figure*}[!ht]
\centering
\includegraphics[width=\textwidth]{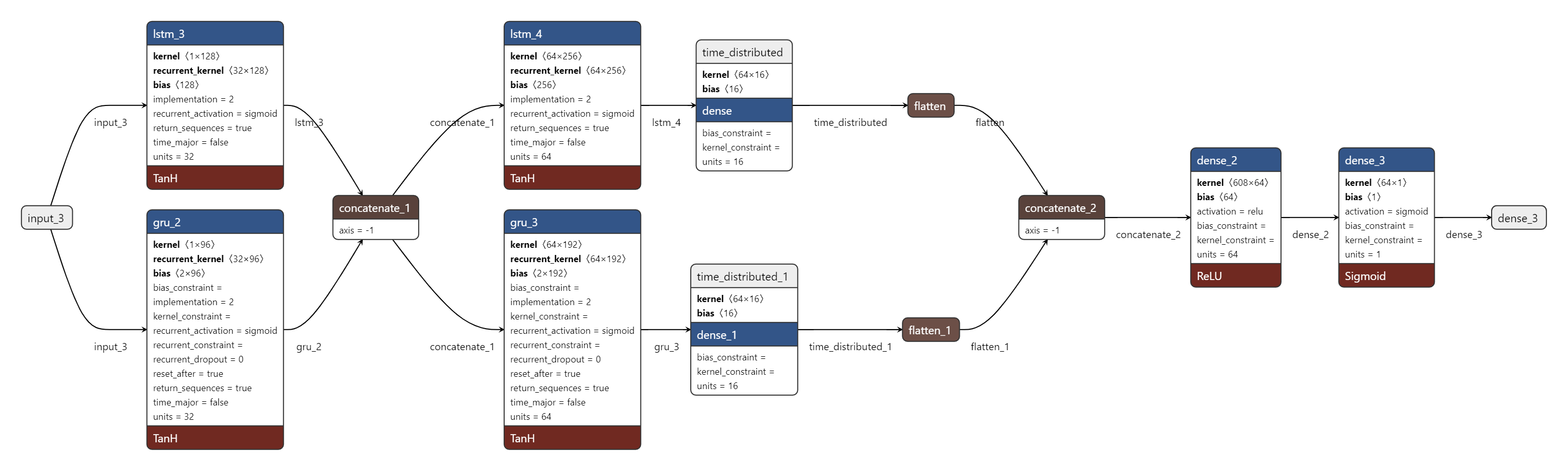}
\caption{SSLR and TL based LSTM-GRU Model Architecture and parameter Settings (\textit{SSRepL-ADHD})}
\label{model} 
\end{figure*}

\subsection{Model Selection}
In this study, different AI techniques are used to improve the process of representation learning. The RF algorithm was used due to its robustness in dealing with complex datasets and its ability to provide effective insights into the importance of features. In addition, SSRepL and TL with LSTM-GRU layers were used in this work to utilize the power of recurrent neural networks and capture temporal dependencies in the data. In addition, an optimized strategy was applied using an LSSRepL DNN to ensure efficiency in processing sequences while balancing model complexity and computational resources. This combination aims to utilize the strengths of each method for thorough and effective learning.

\subsubsection{Random Forest}
The RF model is a versatile and robust ML technique that can be used for classification and regression. The RF model was created with the scikit-learn library. The parameters provided are 'n\_estimators=100', which indicates that 100 decision trees are formed in the forest, and 'random\_state=42', which ensures that the same results are returned when the model is run multiple times with the same data. These parameters contribute to stability and dependability by utilizing an ensemble of decision trees to produce accurate predictions while controlling the unpredictability of the tree-building process.

\subsubsection{Lightweight SSRepL based DNN model (LSSRepL-DNN)}
The LSSRepL-DNN model was developed for learning and downstream binary classification. The model uses a learned representation layer that was initially trained with shallow Long Short-Term Memory (LSTM) layers. The representation layer is frozen to preserve the learned features for future training. A dense layer with ReLU activation is used to create the binary classification model, followed by a final classification layer with a sigmoid activation function for binary classification. A DNN model encapsulates the overall architecture, and the model is built using the Adam optimizer and the binary cross entropy loss function. The model is fed with input data during training, and the learned representations from the frozen layer are used to perform binary classification. The accuracy and loss measures are tracked during the training procedure. The model is trained over 10 epochs with a batch size of 64 using the training data and verified with a separate test set. The model provides information about the layer structure, the number of parameters and the trainable and non-trainable parameters. The described DNN architecture with fewer layers aims to achieve a balance between model complexity and efficiency that enables effective representation learning and subsequent binary classification tasks in this paper.
\subsubsection{SSRepL and TL-based LSTM-GRU Model (SSRepL-ADHD)}
SSRepL-ADHD uses a combination of GRU and LSTM layers to learn representations. To capture subtle temporal correlations in the input data, the model architecture in Fig. \ref{model} consists of many LSTM and GRU layers. These layers are linked together to create a unified representation that is then further processed by LSTM and GRU layers. The model contains fully linked layers for self-supervised learning, which are then flattened and concatenated to produce a consolidated representation. The final layers help with the downstream binary classification task, including fully linked layers for transfer learning and a binary classification output layer. The Adam optimizer, binary cross-entropy loss function, and accuracy as an evaluation metric are used to build the model. The model is trained using 40 epochs with 32 batch sizes of training and validation data.

\subsection{Proposed Algorithm}
Algorithm \ref{transfer_learning_algorithm} explains the proposed model for detecting ADHD patients. The method begins by loading $.mat$ files, labelled as f1, f2, f3, and f4. These files are then utilized to extract relevant data, yielding feature vectors $\mathbf{X}$ and labels $\mathbf{y}$. This extracted data is used to create the final dataset, indicated as $D$. An extensive data preparation phase is applied to the input $D$. The data is cleaned to remove any inconsistencies or noise, yielding $\mathbf{X}_{\text{clean}}$. Following that, label encoding is used to express categorical characteristics, resulting in $\mathbf{X}_{\text{encoded}}$. Finally, standard scaling is used to normalize the data, yielding $\mathbf{X}_{\text{scaled}}$. These steps are essential for preparing the dataset for good model training. Recognizing the dataset's imbalance, the algorithm uses the SMOTE \cite{chawla2002smote} to build a balanced dataset. This procedure yields $\mathbf{X}_{\text{balanced}}$ and $\mathbf{y}_{\text{balanced}}$, guaranteeing that subsequent models are trained on a dataset that represents both classes. After that, the balanced dataset is divided into training and testing sets. The training set, denoted by $\mathbf{X}_{\text{train}}$ and $\mathbf{y}_{\text{train}}$, accounts for 70\% of the balanced data, while the testing set, designated by $\mathbf{X}_{\text{test}}$ and $\mathbf{y}_{\text{test}}$, accounts for the remaining 30\%.

In the following phase, a transfer learning model for the analysis of EEG signals is developed. The model architecture includes LSTM and GRU layers as well as transfer learning approaches to improve the prediction results. The model receives $\mathbf{X}_{\text{train}}$ as input and is trained over 40 epochs with a batch size of 32. The transfer learning model is trained with the training sets $\mathbf{X}_{\text{train}}$ and $\mathbf{y}_{\text{train}}$. The training procedure fine-tunes the parameters of the model so that it can detect the underlying patterns in the EEG signals that indicate ADHD. The next step is to evaluate the performance of the trained model using the testing sets $\mathbf{X}_{\text{test}}$ and $\mathbf{y}_{\text{test}}$. Accuracy, precision, recall, F1 score, and the confusion matrix are among the evaluation metrics that provide a complete assessment of the model's capacity to predict ADHD. The whole method is mentioned and integrated into Algorithm \ref{transfer_learning_algorithm} and provides a step-by-step guide of the proposed framework for ADHD detection using EEG signal analysis.

\begin{algorithm}[!ht]
\caption{SSRepL-ADHD model}
\label{transfer_learning_algorithm}
\begin{algorithmic}[1]
\STATE \textbf{Input:} Data $D$ with features $\mathbf{X}$ and labels $\mathbf{y}$
\STATE \textbf{Output:} Evaluation metrics - Acc, Pre, Rec, F1
\STATE \textbf{Step 1: Load and Extract Data}
\STATE \quad \text{Read and extract .mat Files: } [f1, f2, f3, f4];
\STATE \quad $\mathbf{X}, \mathbf{y} = \text{Extract Data}(f1, f2, f3, f4)$;
\STATE \quad $\text{Final Prepared Dataset: } D = \text{Create\_DB};(\mathbf{X}, \mathbf{y})$
\STATE \textbf{Step 2: Data Preprocessing}
\STATE \quad $X_{clean} = Data(X)$;
\STATE \quad $X_{encoded} = LE(X_{clean})$;
\STATE \quad $X_{scaled} = SC(X_{encoded})$;
\STATE \textbf{Step 3: Train-Test Split}
\STATE \quad $\mathbf{X}_{\text{train}}, \mathbf{y}_{\text{train}} = \text{Split}(\mathbf{X}, \mathbf{y}, 0.7)$;
\STATE \quad $\mathbf{X}_{\text{test}}, \mathbf{y}_{\text{test}} = \text{Split}(\mathbf{X}, \mathbf{y}, 0.3)$;
\STATE \quad \textbf{Step 4: Data Balancing}
\STATE \quad $X_{\text{train}}, X_{\text{train}} = \text{SMOTE}(X-train, y-train)$;
\STATE \textbf{Step 5: SSRepL-ADHD Learning Model}
\STATE \text{Model Definition:}
\STATE \quad $\text{Input: } \text{Input}(\mathbf{X}[1], 1)$;
\STATE \quad $\text{LSTM1: } \text{LSTM}_{64} = \text{LSTM}(64, \text{True})$;
\STATE \quad $\text{GRU1: } \text{GRU}_{64} = \text{GRU}(64, \text{True})$;
\STATE \quad $\text{Concat: } \text{Concat} = \text{Concat.}([\text{LSTM}_{64}, \text{GRU}_{64}])$;
\STATE \quad $\text{LSTM2: } \text{LSTM}_{32} = \text{LSTM}(32, \text{True})(\text{Concat})$;
\STATE \quad $\text{GRU2: } \text{GRU}_{32} = \text{GRU}(32, \text{True})(\text{Concat})$;
\STATE \quad $\text{FC}_{\text{LSTM}} = \text{TimeDistributed}(\text{Dense}_{16})(\text{LSTM}_{32})$;
\STATE \quad $\text{FC}_{\text{GRU}} = \text{TimeDistributed}(\text{Dense}_{16})(\text{GRU}_{32})$;
\STATE \quad $ \text{Flatten}_{\text{LSTM}} = \text{Flatten}(\text{FC}_{\text{LSTM}})$;
\STATE \quad $\text{Flatten}_{\text{GRU}} = \text{Flatten}(\text{FC}_{\text{GRU}}) $;
\STATE \quad $\text{Combined} = \text{Concat.}([\text{Flatten}_{\text{LSTM}}, \text{Flatten}_{\text{GRU}}])$;
\STATE \quad $\text{FC}_{\text{Transfer}} = \text{Dense}_{64}(\text{ReLU})(\text{Combined}) $;
\STATE \quad $Output= \text{Dense}_{1}(\text{Sigmoid})(\text{FC}_{\text{Transfer}})$;
\STATE \textbf{Step 6: Model Training}
\STATE \quad ${Model} = \text{Train\_Model}(\text{Model}, \mathbf{X}_{\text{train}}, \mathbf{y}_{\text{train}})$;
\STATE \textbf{Step 7: Model Evaluation}
\STATE \quad \text{Evaluation Metrics:} \text{Acc, Pre, Rec, F1};
\STATE \quad $\text{EvaluateModel}(\text{Model}, \mathbf{X}_{\text{test}}, \mathbf{y}_{\text{test}})$;
\STATE \quad $\text{Confusion Matrix}(\text{Model}, \mathbf{X}_{\text{test}}, \mathbf{y}_{\text{test}})$;
 \end{algorithmic}
\end{algorithm}

\section{Experimental Results and Discussion}\label{ev}
The experiments use RF, an LSSRepL-based DNN model and our proposed SSRepL- and TL-based LSTM-GRU sequential model. Accuracy, precision, recall and f-score are performance evaluation measures used to assess the ability of the models to achieve optimal classification performance. The ADHD dataset is used for the experimental analysis. The evaluation measures, which are crucial for the examination of model performance, are used to assess the quality of the machine learning models. The data is divided into two sets: Training and Testing, where 70\% is used for training and 30\% is used for testing.

\subsubsection{Random Forest Results}
\begin{figure*}[!ht]
\centering
\subfloat[Confusion Matrix of Random Forest Model \label{ref_conf}]{ \includegraphics[width=0.33\textwidth]{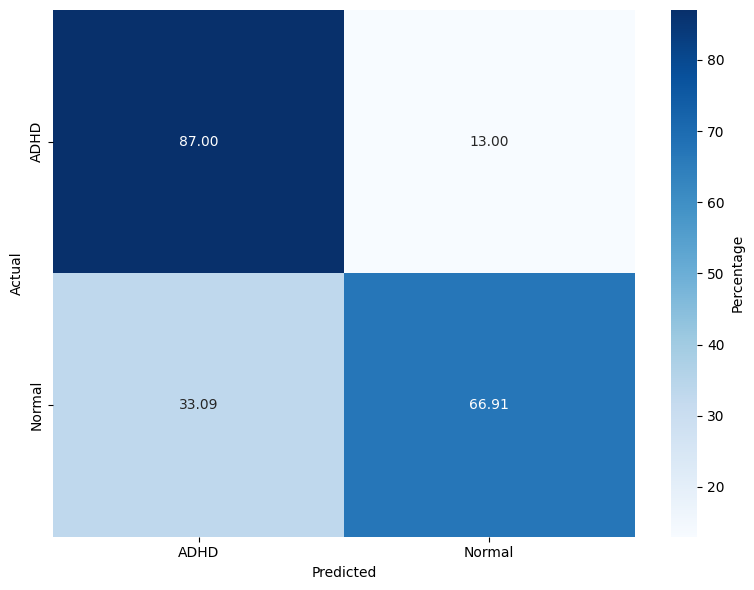}}
\subfloat[Confusion Matrix of DNN Model \label{dl_conf}] {
\includegraphics[width=0.33\textwidth]{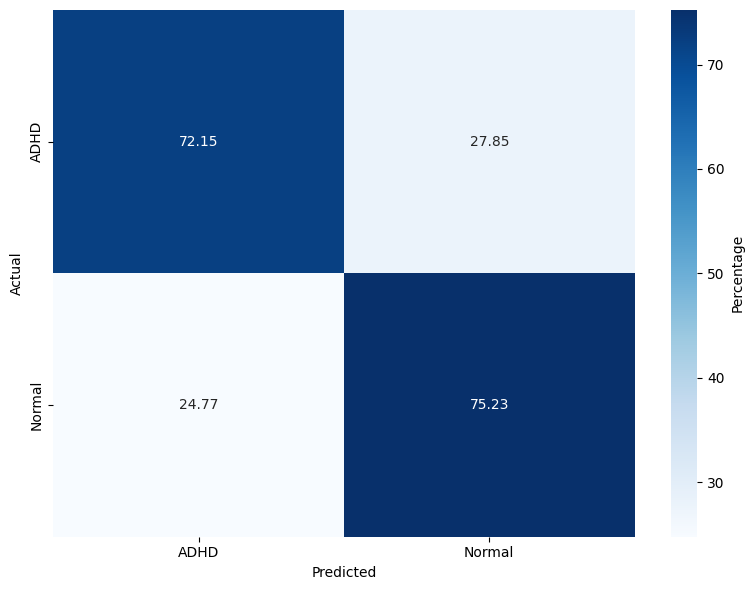}}
\subfloat[Confusion Matrix of \textit{SSRepL-ADHD} model\label{tl_conf}] {
\includegraphics[width=0.33\textwidth]{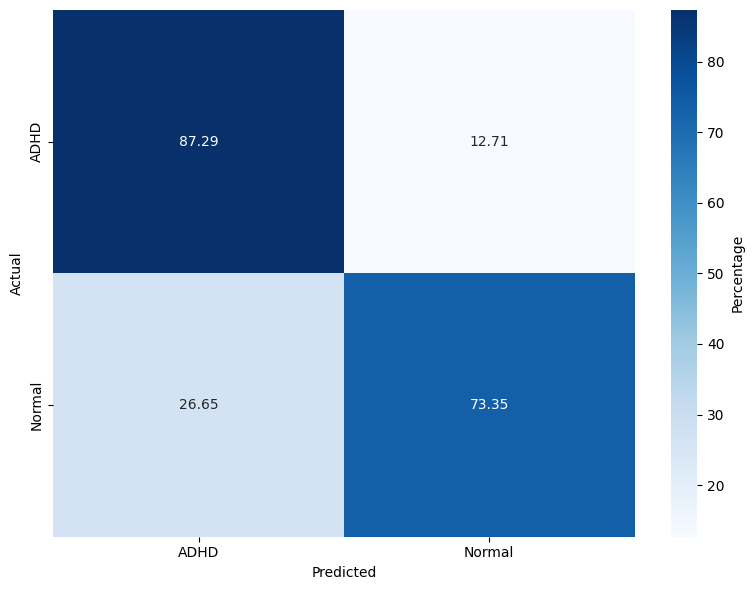}}
 \caption{Confusion Matrix of Classification Models}
\label{cfimagess}
\end{figure*}

TABLE \ref{results} shows the experimental results of the proposed methodology and highlights the performance metrics for different models. The accuracy, precision, recall, and F1-score of the RF model are all 78.01\%, indicating consistent performance on all criteria. Fig. \ref{cfimagess}(a) shows the confusion matrix of the RF model. For the ADHD class, the model correctly categorized 87\% of cases as ADHD (True Positives), while it misidentified 13\% of cases as Normal (False Negatives). In contrast, the model correctly classified 66.91\% instances as Normal (True Negatives), while it incorrectly classified 33.09\% of examples in the Normal class as ADHD (False Positives). The confusion matrix essentially provides a detailed summary of the model's classification performance, detailing the correctly and incorrectly predicted instances for each class, allowing for a more nuanced assessment of the model's strengths and weaknesses.

\begin{table*}[!ht]
\caption{Proposed Methodology Experimental Results (Weighted)}
\label{results}
\centering
\begin{tabular}{|l|l|l|l|l|}
\hline
\textbf{Models} & \textbf{Accuracy(\%)} & \textbf{Precision(\%)} & \textbf{Recall(\%)} & \textbf{F1-score(\%)} \\ \hline
RF & 78.01 & 78.01 & 78.01 & 78.01 \\ \hline
Lightweight SSRepL based DNN Model & 73.67 & 74.05 & 74.05 & 74.05 \\ \hline
\textit{SSRepL-ADHD} Model & 81.11 & 81.10 & 81.10 & 81.10 \\ \hline
\end{tabular}
\end{table*}

\subsubsection{Lightweight TL-SSLR based DNN Results}
The Deep Neural Network (DNN) with fewer layers achieves slightly lower metrics, achieving 73.67\% accuracy and 74.05\% precision, recall, and F1-score. According to Fig. \ref{imagess1}(a) and Fig. \ref{imagess1}(b), the DNN is trained with fewer layers during ten epochs and the evolution of key metrics is tracked. The model has an initial loss of 0.6390 and an accuracy of 63.52\% in the first epoch, while the validation set has a loss of 0.5991 and an accuracy of 68.92\%. As training progresses, there is a visible improvement in performance, with the loss gradually decreasing. By the ninth epoch, the model achieves a training loss of 0.5341 and a training accuracy of 73.95\%. The validation set, on the other hand, shows a modest increase in loss (0.5389) and a slight decrease in accuracy (73.67\%), indicating a potential risk of overfitting. These results represent an iterative change in the weights and biases of the model during training to achieve optimal prediction performance on the training data. It is crucial to properly evaluate the validation metrics as they provide information about the generalization of the model to new data and potential symptoms of overfitting or underfitting.

The confusion matrix for the DNN model with fewer layers, shown in Fig. \ref{cfimagess}(b), summarizes the classification performance of the model for two classes, ADHD and Normal. The model accurately identifies 72.15\% of instances as ADHD (True Positives) and 27.85\% of instances incorrectly as Normal (False Negatives) for the ADHD class. For the Normal class, on the other hand, the model correctly identifies 75.23\% of instances as Normal (True Negatives) but incorrectly classifies 24.77\% of instances as ADHD (False Positives). This matrix provides a detailed overview of the model's capacity to make correct and incorrect predictions for each class, allowing a more efficient evaluation of its performance in differentiating between ADHD and Normal instances.

\subsubsection{SSRepL-ADHD Model Results}
\begin{figure*}[!ht]
\centering
\subfloat[Training and Validation Accuracy of DNN Model \label{dnn_acc}]{
 \includegraphics[width=0.5\textwidth]{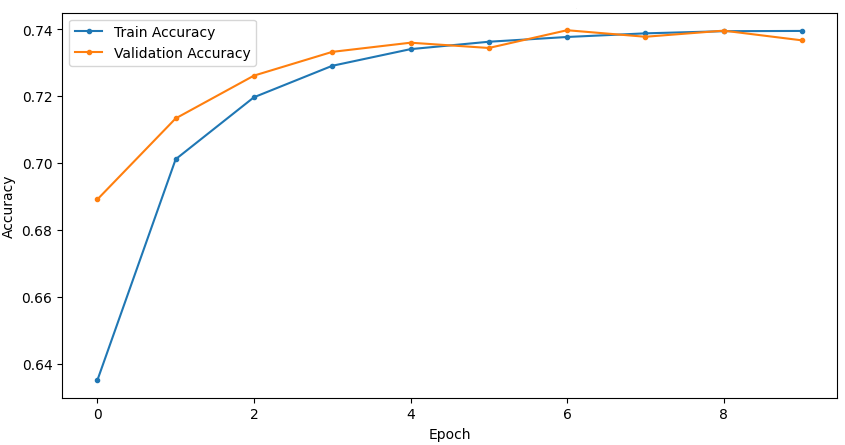}}
\subfloat[Training and Validation Loss of DNN Model \label{dnn_loss}] {
\includegraphics[width=0.5\textwidth]{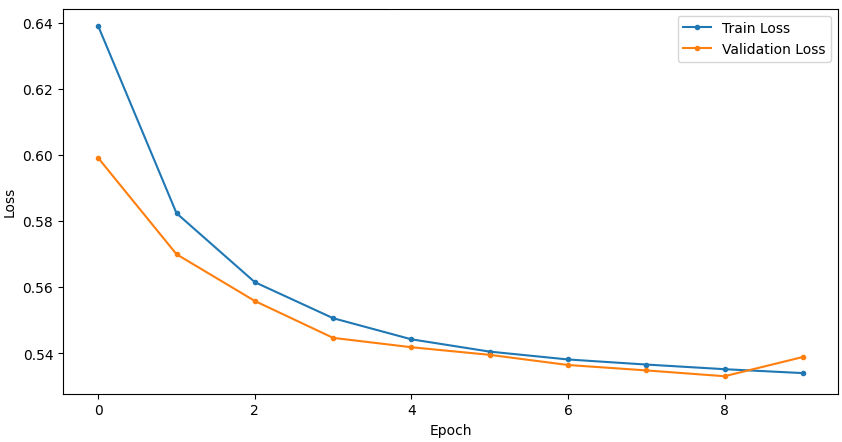}}\\
\subfloat[Training and Validation Accuracy of SSRepL-ADHD model \label{tl_acc}] {
\includegraphics[width=0.5\textwidth]{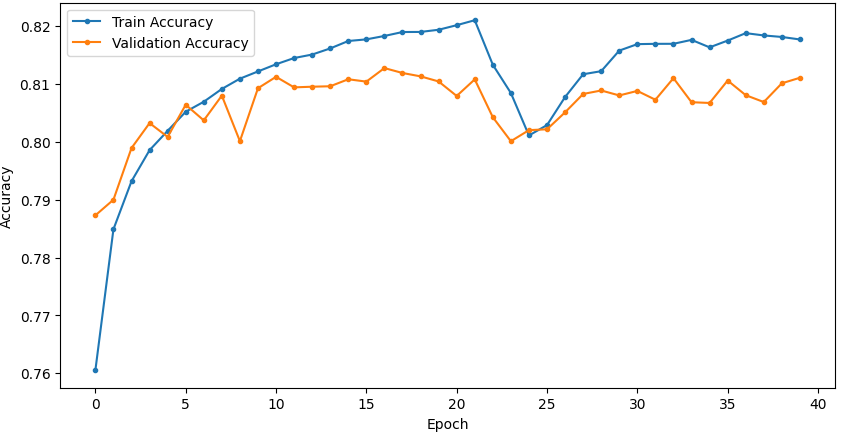}}
\subfloat[Training and Validation Loss of SSRepL-ADHD model\label{tl_loss}] {
\includegraphics[width=0.5\textwidth]{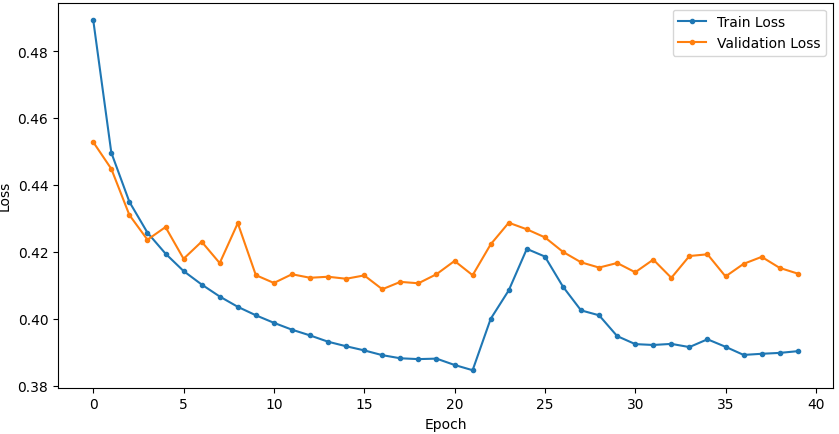}}
 \caption{Graphical Results of Training and Validation}
\label{imagess1}
\end{figure*}

On the other hand, the SSRepL-ADHD model outperforms the other models with accuracy, precision, recall, and an F1-score of 81.11\%. These results show that the proposed model achieves higher classification performance compared to the RF and DNN models on many evaluation metrics. According to Figs. \ref{imagess1}(c) and \ref{imagess1}(d), the transfer learning model containing LSTM and GRU layers is trained over 40 epochs, each with its own loss and accuracy metrics. The model starts with a training loss of 0.4894 and an accuracy of 76.05\% in the first epoch, while the validation set shows a loss of 0.4529 and an accuracy of 78.73\%. The model's ability to learn and generalize patterns from data is evident in a significant improvement in the training and validation measures as training progresses. The designed model achieves a training loss of 0.3903 and a training accuracy of 81.77\% by the last epoch. The validation set yields a loss of 0.4135 and an accuracy of 81.11\%. These results indicate that the knowledge from the pre-trained LSTM and GRU layers was successfully learned and transferred, resulting in an excellent model for categorizing ADHD and Normal instances. The lengthy training time of around 22 hours and 57 minutes illustrates the complexity of the model and the intensive learning process required for maximum performance.

As shown in Fig. \ref{cfimagess}(c), the confusion matrix for the SSRepL-ADHD model gives a deep insight into the classification performance of the model. The model accurately classifies 87.29\% of cases as ADHD (true positive), but 12.71\% of cases are mislabeled as Normal (false negative) in the context of ADHD categorization. In contrast, for Normal cases, the model correctly predicts 73.35\% of cases (true negative) but misclassifies 26.65\% of cases as ADHD (false positive). The confusion matrix provides a detailed evaluation of the strengths and limitations of the model and helps to interpret its classification capabilities.
.

\subsection{Discussion}
As ADHD is the most prevalent disorder in children, early detection will help prevent future difficulties \cite{mohammadi2016eeg}. In this paper, ADHD patients were thoroughly identified by analyzing EEG signals using a number of machine and deep learning approaches such as RF, the LSSRepL-DNN model and the proposed SSRepL-ADHD sequential model. The results of the experiments are promising, with accuracy, precision, recall and F1-score serving as crucial criteria for performance evaluation. The paper aims to exploit the strengths of each approach for effective representation learning by leveraging the robustness of RF on complex datasets and feature importance insights, transfer learning with LSTM and GRU layers to capture temporal dependencies, and a streamlined DNN with fewer layers for efficient sequence processing. A systematic classification approach includes data preparation, preprocessing, handling imbalanced datasets and model implementation. The study emphasizes the importance of certain features and evaluates the performance of the model using a dataset collected from EEG signals.

Despite the positive results, there are also difficulties. The imbalance of the dataset requires special attention, and approaches such as data balancing are used to reduce bias. The selection of AI approaches requires a thorough analysis of their advantages and disadvantages. Moreover, while the study recognizes the importance of each attribute, it faces the problem of selecting the most relevant ones. The experimental results presented in TABLE \ref{results} show the usefulness of the models, with TL with LSTM and GRU layers achieving the best accuracy of 81.11\%. The RF and LSSRepL-DNN models perform well with an accuracy of 78.01\% and 73.67\%, respectively. The indicators of precision, recall, and F1-score are consistent with the overall accuracy trends. Finally, using multiple AI algorithms, this study provides valuable insights into the classification of ADHD based on EEG signals. The difficulties encountered, including the imbalance of the dataset and feature selection, demonstrate the importance of the methodology. The results confirm the usefulness of the developed model while admitting the continuous effort of refining model performance and addressing intrinsic obstacles in EEG signal processing for ADHD classification.

\section{Conclusion} \label{conclusion}
This study provided a detailed analysis of ADHD classification by analyzing EEG data utilizing a variety of machine and deep learning techniques. The results show that these models can reliably diagnose ADHD patients, with each model having a unique accuracy, precision, recall and F1 score. The RF model achieved an accuracy of 78.01\%, demonstrating its robustness in dealing with complicated datasets and providing important insights into feature relevance. The LSSRepL-DNN model, designed for efficient processing of sequences, achieved a competitive accuracy of 73.67\%, highlighting its utility in balancing model complexity and computational economy. The SSRepL-ADHD model achieved the greatest accuracy of 81.11\%, utilizing the ability of recurrent neural networks to capture data temporal connections. These numerical results show that the proposed SSRepL-ADHD model outperforms the other strategies, highlighting its potential to improve ADHD classification accuracy. The precision, recall, and F1-score measures correlate with the overall accuracy trends, validating the models' success across various performance evaluation parameters. Although the results are encouraging, it is important to recognize the difficulties encountered during the research, such as the imbalance of the dataset and the selection of key attributes. Overcoming these difficulties requires continuous improvement of the methodology and the search for innovative solutions. Despite these obstacles, the results provide valuable insights into the categorization of ADHD based on EEG signals.

Future studies should focus on overcoming these limitations by investigating new strategies for dealing with imbalanced datasets and improving feature selection processes. Additional data modalities or better signal processing methods could be incorporated into future studies to improve the ability of models to detect subtle patterns suggestive of ADHD. In addition, investigating interpretation methods and model explainability could help build confidence in the model's predictions, especially in other clinical applications.

\ifCLASSOPTIONcaptionsoff
\newpage
\fi

\bibliographystyle{IEEEtran}
\bibliography{ref}
\end{document}